\documentclass{article}

\usepackage[preprint]{neurips_2026}
\usepackage{subcaption}
\usepackage[utf8]{inputenc} 
\usepackage[T1]{fontenc}    
\usepackage{hyperref}       
\usepackage{url}            
\usepackage{booktabs}       
\usepackage{amsfonts}       
\usepackage{nicefrac}       
\usepackage{microtype}      
\usepackage{xcolor}         
\usepackage{amsmath} 
\usepackage{multirow}
\usepackage[table]{xcolor}
\usepackage{graphicx}
\usepackage{adjustbox}
\usepackage{array}
\usepackage{algorithm}
\usepackage{algorithmic}
\usepackage{amsmath}
\definecolor{attackred}{RGB}{165,65,65}   
\definecolor{dropblue}{RGB}{70,105,150}   
\definecolor{cleangray}{RGB}{45,45,45}    
\definecolor{summarygray}{RGB}{238,238,238}
\newcommand{\ca}[2]{%
  \textcolor{cleangray}{#1}/\textbf{\textcolor{attackred}{#2}}%
}

\newcommand{\avgdrop}[2]{%
  \textbf{\textcolor{dropblue}{#1}}%
  {\scriptsize\textcolor{dropblue}{\, (#2\%)}}%
}
\usepackage{pifont}
\newcommand{\cmark}{\ding{51}}
\newcommand{\xmark}{\ding{55}}
\title{Mistletoe: Stealthy Acceleration-Collapse Attacks \\on Speculative Decoding}

\author{
 \textbf{Shuoyang Sun\textsuperscript{1}$^{\dag}$},
 \textbf{Chang Dai\textsuperscript{2}$^{\dag}$},
 \textbf{Hao Fang\textsuperscript{3}},
 \textbf{Kuofeng Gao\textsuperscript{3}},
 \textbf{Xinhao Zhong\textsuperscript{1}},
\\
 \textbf{Yi Sun\textsuperscript{1}},
 \textbf{Fan Mo\textsuperscript{4}},
 \textbf{Shu-Tao Xia\textsuperscript{3}},
 \textbf{Bin Chen\textsuperscript{1}$^{*}$}
\\
 \textsuperscript{1}Harbin Institute of Technology, Shenzhen
\\
 \textsuperscript{2}South China University of Technology
\\
 \textsuperscript{3}Tsinghua Shenzhen International Graduate School, Tsinghua University
\\
 \textsuperscript{4}Huawei Technology
}

\begin{document}
\maketitle
{
  \renewcommand{\thefootnote}{} 
  \footnotetext{$^\dagger$ Equal contribution.}
  \footnotetext{$^\ast$ Corresponding author.}
}

\maketitle

\begin{abstract}
Speculative decoding has become a widely adopted technique for accelerating large language model (LLM) inference by drafting multiple candidate tokens and verifying them with a target model in parallel. Its efficiency, however, critically depends on the average accepted length $\tau$, i.e., how many draft tokens survive each verification step. In this work, we identify a new mechanism-level vulnerability in model-based speculative decoding: the drafter is trained to approximate the target model distribution, but this approximation is inevitably imperfect. Such a drafter--target mismatch creates a hidden attack surface where small perturbations can preserve the target model's visible behavior while substantially reducing draft-token acceptability.
We propose \textsc{Mistletoe}, a stealthy acceleration-collapse attack against speculative decoding. \textsc{Mistletoe} directly targets the acceptance mechanism of speculative decoding. It jointly optimizes a degradation objective that decreases drafter--target agreement and a semantic-preservation objective that constrains the target model's output distribution. To resolve the conflict between these objectives, we introduce a null-space projection mechanism, where degradation gradients are projected away from the local semantic-preserving direction, suppressing draft acceptance while minimizing semantic drift. Experiments on various speculative decoding systems show that \textsc{Mistletoe} substantially reduces average accepted length $\tau$, collapses speedup, and lowers averaged token throughput, while preserving output quality and perplexity. 
Our work highlights that speculative decoding introduces a mechanism-level attack surface beyond existing output robustness, calling for more robust designs of LLM acceleration systems.
\end{abstract}

\section{Introduction}

Large language models (LLMs) have achieved remarkable capabilities across open-ended generation, reasoning, and interactive assistance~\citep{grattafiori2024llama, liu2024deepseek, yang2025qwen3}. 
However, autoregressive decoding remains inherently sequential, as each generated token requires a separate target-model invocation conditioned on the preceding context. 
Speculative decoding mitigates this bottleneck through a draft-then-verify paradigm: a lightweight drafter proposes candidate continuations, and the target model verifies them in parallel~\citep{leviathan2023fast, chen2023accelerating}. 
By accepting multiple draft tokens in one target-model forward pass, speculative decoding can accelerate generation while preserving the target model's output distribution under standard verification rules. 
Thus, the practical efficiency of speculative decoding depends not merely on how many tokens are drafted, but on how many of them are accepted by the target verifier. 
The average accepted length \(\tau\) therefore captures a central mechanism behind speculative acceleration.

Recent speculative decoding systems explicitly optimize this mechanism by improving drafter--target agreement. 
They introduce auxiliary prediction heads, target-model features, dynamic draft trees, fused intermediate representations, or shared computation to make draft proposals more acceptable to the target verifier~\citep{cai2024medusa, li2024eagle, li2024eagle2, ankner2024hydra, li2025eagle}. 
Alignment-oriented work further shows that mitigating token and feature misalignment improves draft-token acceptance, accepted length, and speedup~\citep{hu2025griffin}. 
These advances reveal that acceptance is not a secondary implementation detail, but a critical foundation that underpins the efficiency of speculative decoding. 
While residual drafter--target mismatch is usually treated as an efficiency bottleneck to be reduced, we show that it can also become an attack surface.

Existing safety-oriented studies on speculative decoding mainly examine privacy leakage or generated-content safety. 
Input-dependent speculation patterns may create side channels that leak private information~\citep{wei2024speculation}, while safety-aware decoding methods use auxiliary or small expert models to improve output safety~\citep{wang2025speculative, wang2025secdecoding}. 
In this paper, we raise a largely unexplored mechanism-level security question: 
\textit{Can the draft-verification pathway itself be adversarially degraded while the final response remains visibly normal?}
If a small perturbation preserves the target model's response distribution while causing drafter proposals to diverge from the target verifier, drafted tokens will be repeatedly rejected during verification. 
Consequently, $\tau$ collapses, speedup disappears, and averaged token throughput decreases. 
We define this failure mode as an \emph{acceleration-collapse attack}, which disables the mechanism that makes generation fast while preserving generated-content quality.

\begin{figure}[t]
    \centering
    \includegraphics[width=\linewidth]{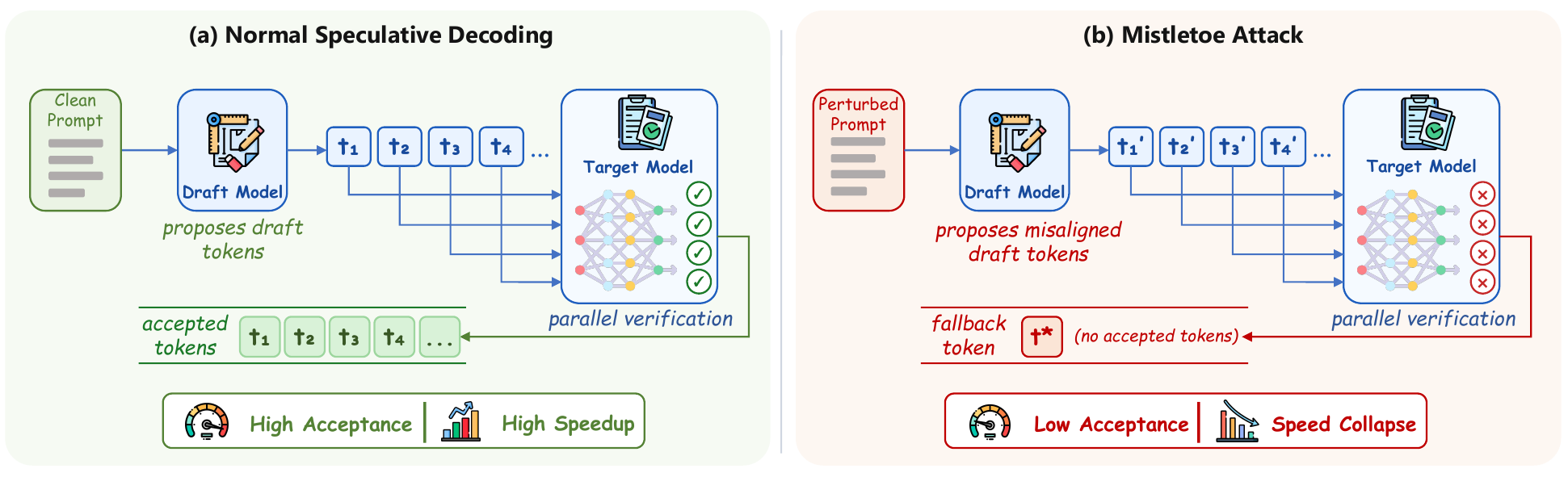}
    \caption{
    Illustration of acceptance collapse under \textsc{Mistletoe}.
    In normal speculative decoding, the target model accepts many draft tokens per verification step, yielding high acceptance and high speedup.
    When speculative decoding is attacked by \textsc{Mistletoe}, the final response semantics remain preserved, but misaligned draft tokens are rejected by the target verifier, forcing fallback generation from target logits and collapsing the average accepted length $\tau$ and speedup.
    }
    \vspace{-1em}
    \label{fig:intro_comparison}
\end{figure}

Based on these insights, we propose \textsc{Mistletoe}, a stealthy acceleration-collapse attack against speculative decoding. As illustrated in Figure~\ref{fig:intro_comparison}, normal speculative decoding commits multiple accepted draft tokens per verification step, whereas decoding under \textsc{Mistletoe} frequently falls back to target-generated tokens. This definition reflects the attack's parasitic nature of the attack: it remains unobtrusive at the output level while draining the efficiency benefit of the host decoding pipeline. Unlike content-level attacks that aim to alter the generated response, \textsc{Mistletoe} targets the core verification-and-acceptance mechanism that makes speculative decoding fast. It increases target-side surprisal of drafter-proposed tokens to reduce their acceptability, while constraining the target model's output distribution to preserve response quality.

To reach the attack goal, a key challenge is the optimization conflict of drafter--target alignment. 
Draft tokens are designed to approximate the target model's high-probability continuations; therefore, perturbations that reduce draft-token acceptability can also disturb the target model's own output distribution. 
To address this conflict, \textsc{Mistletoe} restricts the rejection direction to the local null space of the semantic-preservation constraint~\citep{fang2024alphaedit}. 
This projection encourages updates that increase rejection pressure while limiting semantic drift. 
A KL-threshold filter further vetoes discrete suffix candidates whose target-distribution drift exceeds a preset bound.

Experiments on representative speculative decoding systems show that \textsc{Mistletoe} substantially reduces $\tau$, speedup, and averaged token throughput, while preserving output quality and perplexity. 
These results demonstrate that speculative decoding can be vulnerable even when user-facing textual outputs appear normal, highlighting the need for robust and security-aware acceleration mechanisms.

In summary, our contributions are threefold:
\begin{itemize}
    \item We identify \emph{acceptance collapse} as a mechanism-level threat to speculative decoding, where adversarially amplified drafter--target mismatch reduces the average accepted length $\tau$ while leaving user-facing outputs largely preserved.
    \item We propose \textsc{Mistletoe}, a stealthy acceleration-collapse attack that degrades the verification-and-acceptance pathway rather than directly corrupting generated content.
    \item We develop a null-space projected optimization method with KL-threshold filtering to reduce draft-token acceptability while suppressing semantic drift, and empirically demonstrate substantial degradation of speculative decoding efficiency across representative systems.
\end{itemize}

\section{Related Work}

\subsection{Speculative Decoding for Efficient Inference}

Speculative decoding accelerates autoregressive generation through a draft-then-verify paradigm, where a lightweight drafter proposes candidate continuations and the target model verifies them in parallel~\citep{leviathan2023fast, chen2023accelerating}. 
Its efficiency depends on accepting multiple draft tokens per target-model forward pass, making the average accepted length central to realized speedup.

Existing work mainly improves this mechanism by enhancing draft quality or verification efficiency through multiple decoding heads, target-model features, dynamic draft trees, sequentially dependent draft heads, and multi-layer feature fusion~\citep{cai2024medusa, li2024eagle, li2024eagle2, ankner2024hydra, li2025eagle}. 
Recent surveys further cover independent-drafter, retrieval- or n-gram-based, model-free, self-speculative, and draft-head-based variants~\citep{hu2025speculative}. 
Despite architectural differences, these methods share a common objective: increasing drafter--target agreement so that more candidate tokens are accepted per verification step. 
GRIFFIN further highlights this dependency by identifying token and feature misalignment as a bottleneck for draft-token acceptance and improving accepted length by mitigating such misalignment~\citep{hu2025griffin}. 
Together, these studies establish drafter--target agreement as a central determinant of speculative decoding efficiency.

\subsection{Safety and Robustness of Speculative Decoding}

As speculative decoding becomes increasingly relevant to efficient LLM deployment, recent studies have examined its safety implications. 
One line studies privacy leakage through input-dependent speculation patterns, where observable decoding behaviors may create side channels~\citep{wei2024speculation}. 
Another leverages speculative or auxiliary-model decoding to improve output safety, such as detecting jailbreak risks or constructing token-level safety signals for safer generation~\citep{wang2025speculative, wang2025secdecoding}. 
These works provide valuable insights, but they focus on privacy leakage or generated-content safety and leave the robustness of the acceleration mechanism itself largely unexplored.

In contrast, we take a mechanism-level perspective: the drafter--target agreement that enables acceleration also defines a fragile boundary of speculative decoding. 
\textsc{Mistletoe} shows that adversarial perturbations can amplify drafter--target mismatch, collapsing draft-token acceptance while keeping final response behavior largely preserved. 
This exposes a performance-robustness threat to speculative decoding, complementing prior studies on privacy and output-level safety.

\section{Preliminaries}
\label{sec:prelim_formulation}

\subsection{Speculative Decoding}
\label{sec:prelim_sd}

We formalize the draft-then-verify process underlying speculative decoding. 
Let \(M_{\theta}\) denote the target language model, or verifier, and \(D_{\phi}\) denote the drafter. 
Given a prompt \(x\), \(D_{\phi}\) proposes draft tokens and \(M_{\theta}\) verifies them in parallel. 
Let \(t\) index draft-then-verify cycles, and let \(Y^{(t)}\) denote the accepted output prefix before the \(t\)-th cycle. 
Within this cycle, the drafter proposes \(\hat y^{(t)}_1,\ldots,\hat y^{(t)}_K\), where \(K\) is the draft budget and \(i\) indexes the position within the current draft.

We denote the drafter distribution for the \(i\)-th draft token by
\(\rho_{\phi}(\cdot \mid x, Y^{(t)}, \hat y^{(t)}_{<i})\), where \(\hat y^{(t)}_{<i}\) represents previously drafted tokens in the same chain or tree branch. 
The target verifier distribution is denoted by
\(\pi_{\theta}(\cdot \mid x, Y^{(t)}, \hat y^{(t)}_{<i})\). 
This notation abstracts over different implementations: \(\rho_{\phi}\) may be produced by an independent draft model, auxiliary draft heads, or feature-reuse modules. 
In all cases, acceleration is governed by the agreement between \(\rho_{\phi}\) and \(\pi_{\theta}\).

A draft token is likely to be accepted only when the target assigns it probability comparable to the drafter. 
For a chain draft under the standard speculative verification rule~\citep{leviathan2023fast, chen2023accelerating}, the acceptance probability of \(\hat y^{(t)}_i\) is
\begin{equation}
\alpha^{(t)}_{i}
=
\min\!\left(
1,\,
\pi_{\theta}\!\left(\hat y^{(t)}_{i}\mid x, Y^{(t)}, \hat y^{(t)}_{<i}\right)
\big/
\rho_{\phi}\!\left(\hat y^{(t)}_{i}\mid x, Y^{(t)}, \hat y^{(t)}_{<i}\right)
\right).
\label{eq:acceptance_probability}
\end{equation}
If accepted, the draft token is committed to the output prefix; otherwise, the verifier falls back to the target logits and discards the remaining draft tokens along that path. 
Thus, \(\{\alpha_i^{(t)}\}_{i=1}^{K}\) determine how many draft tokens survive verification. 
Let \(a^{(t)}\) be the number of tokens committed in the \(t\)-th cycle, including accepted draft tokens and the target-generated fallback token. 
The average accepted length is
\begin{equation}
\tau
=
\mathbb{E}_{t}
\left[
a^{(t)}
\right].
\label{eq:tau}
\end{equation}
In practice, \(\tau\) is measured as the number of generated tokens divided by the number of target-model forward passes. 
A large \(\tau\) indicates effective amortization of target computation, whereas \(\tau \approx 1\) indicates degeneration toward vanilla autoregressive decoding. 
For tree-based speculative decoding, multiple branches are verified with tree attention, but the same acceptance principle holds.

\subsection{Threat Model and Objective}
\label{sec:threat_objective}

We consider an adversary who aims to degrade speculative acceleration without directly corrupting the final response. 
The target model \(M_{\theta}\) and drafter \(D_{\phi}\) remain fixed. 
The adversary appends a short discrete suffix \(\delta \in \mathcal{V}^{m}\) to the clean prompt \(x\), producing
\begin{equation}
x_{\delta}
=
x
\oplus
\delta ,
\label{eq:adv_prompt}
\end{equation}
where \(\mathcal{V}\) is the vocabulary, \(m\) is the suffix length, and \(\oplus\) denotes concatenation. 
During attack construction, the adversary uses white-box gradients to optimize \(\delta\); at deployment time, the attack only requires submitting \(x_{\delta}\).

Under \(x_{\delta}\), speculative decoding follows the same procedure, but the drafter and verifier distributions become
\(\rho_{\phi}(\cdot \mid x_{\delta}, Y^{(t)}, \hat y^{(t)}_{<i})\) and
\(\pi_{\theta}(\cdot \mid x_{\delta}, Y^{(t)}, \hat y^{(t)}_{<i})\). 
The attack seeks to make draft tokens less acceptable to the target verifier, thereby reducing \(a^{(t)}\), collapsing \(\tau\), and eliminating the speedup benefit. 
Meanwhile, the target model's output distribution should remain close to its clean behavior so that user-facing responses are largely preserved.

We formulate this goal as a constrained discrete optimization problem:
\begin{equation}
\max_{\delta \in \mathcal{V}^{m}}
\quad
\mathcal{L}_{\mathrm{rej}}(x,\delta)
\qquad
\mathrm{s.t.}
\qquad
\mathcal{L}_{\mathrm{sem}}(x,\delta)
\le
\epsilon .
\label{eq:attack_objective}
\end{equation}
Here, \(\mathcal{L}_{\mathrm{rej}}\) measures rejection pressure on drafter-proposed tokens, while \(\mathcal{L}_{\mathrm{sem}}\) defines the semantic-preservation constraint by bounding target-model distributional drift induced by the suffix. 
We instantiate these terms in the next section.

\begin{figure*}[t]
    \centering
    \includegraphics[width=\textwidth]{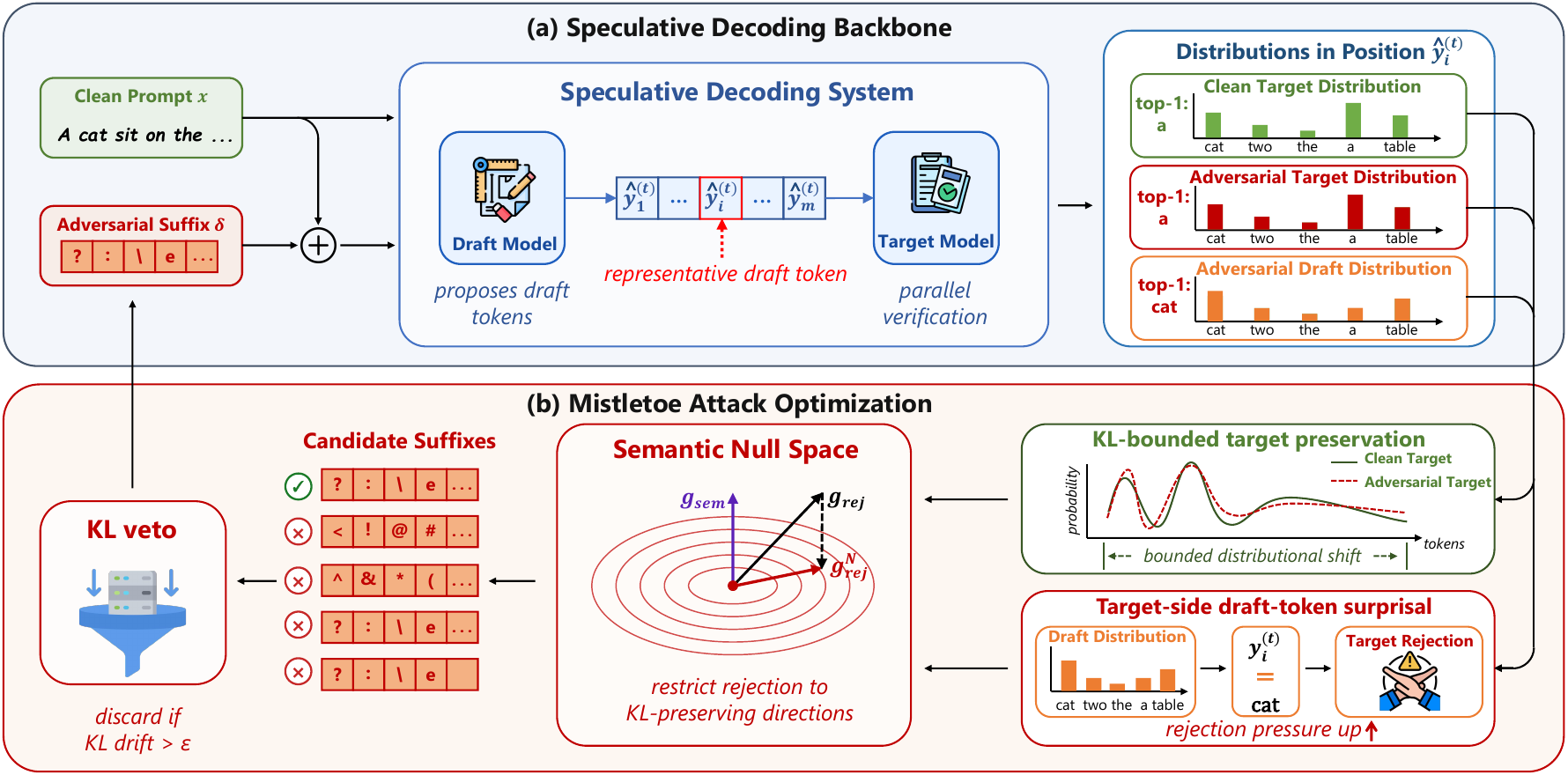}
    \caption{
    Pipeline of \textsc{Mistletoe}. 
    The adversarial suffix \(\delta_k\) is appended to the clean prompt \(x\) and passed through the speculative decoding system. 
    We visualize one representative draft token \(\hat y_i^{(t)}\); in practice, the objectives aggregate over multiple positions. 
    Target-side Draft-Token Surprisal increases rejection pressure by reducing the target verifier's confidence in drafter-proposed tokens, while KL-bounded Target Preservation constrains the adversarial target distribution to remain close to the clean one. 
    The rejection direction is restricted to the semantic null space, producing a feasible update direction for discrete suffix search. 
    A KL-bound veto filters infeasible candidates, and the selected high-surprisal suffix becomes \(\delta_{k+1}\).
    }
    \vspace{-0.5em}
    \label{fig:method_pipeline}
\end{figure*}

\section{Method}

\subsection{Overview}
\label{sec:method_overview}

To collapse speculative acceleration while preserving final-response behavior, \textsc{Mistletoe} optimizes a short discrete suffix appended to the clean prompt, as illustrated in Figure~\ref{fig:method_pipeline}. 
It combines two information-theoretic objectives: Target-side Draft-Token Surprisal increases the mismatch between drafter proposals and the target verifier, reducing draft-token acceptability; KL-bounded Target Preservation constrains the adversarial target distribution to remain close to the clean one. 
Because these objectives can induce entangled gradients, \textsc{Mistletoe} restricts the rejection direction to the local null space of the semantic-preservation constraint, and uses the resulting feasible direction to guide discrete suffix search with a KL-bound veto. 
Together, these components reduce the average accepted length \(\tau\) and speculative speedup while keeping user-facing responses largely preserved.

\subsection{On Acceptance Collapse: Surprisal under KL Constraints}
\label{sec:info_objectives}

As discussed in Section~\ref{sec:prelim_sd}, speculative decoding derives its speedup from the draft-then-verify mechanism, where drafter-proposed tokens are committed only when they remain sufficiently likely under the target verifier. 
This mechanism naturally defines the attack target of \textsc{Mistletoe}: rather than directly corrupting the final response, we aim to reduce the acceptability of draft tokens under the target verifier. 
A direct way to do so is to lower the target probability assigned to drafter-proposed tokens, which decreases their acceptance probability, reduces the committed length \(a^{(t)}\), and ultimately collapses the average accepted length \(\tau\). 
However, this strategy is difficult to optimize safely. 
Speculative decoding is effective precisely because the drafter is designed to approximate the target verifier; hence, accepted draft tokens often lie close to the target model's own high-probability region. 
Aggressively suppressing these tokens can therefore distort the target distribution and expose the attack through semantic drift or abnormal outputs.

To instantiate the constrained objective defined in Section~\ref{sec:threat_objective}, we use two complementary quantities. 
The first increases the target-side surprisal of draft tokens, creating rejection pressure against the acceleration pathway. 
The second bounds the distributional drift of the target verifier, preserving the model's clean response behavior.

\paragraph{Target-side draft-token surprisal.}
For a drafter-proposed token \(\hat y_i^{(t)}\), its surprisal under the adversarial prompt \(x_\delta\) is
\begin{equation}
s_{\theta}\!\left(\hat y_i^{(t)};x_\delta\right)
=
-\log
\pi_{\theta}\!\left(
\hat y_i^{(t)}
\mid
x_{\delta},Y^{(t)},\hat y_{<i}^{(t)}
\right).
\label{eq:draft_surprisal}
\end{equation}
Maximizing this quantity lowers the target verifier's confidence in drafter-proposed tokens, hence increasing rejection pressure. 
Given attacked draft-token positions \(\mathcal{I}\), we define the rejection objective:
\begin{equation}
\mathcal{L}_{\mathrm{rej}}(x,\delta)
=
\frac{1}{|\mathcal{I}|}
\sum_{(t,i)\in\mathcal{I}}
s_{\theta}\!\left(\hat y_i^{(t)};x_\delta\right).
\label{eq:rejection_loss}
\end{equation}

\paragraph{KL-bounded target preservation.}
While \(\mathcal{L}_{\mathrm{rej}}\) targets the acceleration pathway, the semantic constraint requires the target model's own output behavior to remain close to its clean behavior. 
We therefore use the clean target distribution as an information-theoretic reference for the adversarial target distribution. 
Let \(\mathcal{S}\) denote the positions used to estimate distributional drift. 
We define
\begin{equation}
\mathcal{L}_{\mathrm{sem}}(x,\delta)
=
\frac{1}{|\mathcal{S}|}
\sum_{t\in\mathcal{S}}
D_{\mathrm{KL}}
\left(
\pi_{\theta}(\cdot\mid x,Y^{(t)})
\,\middle\|\,
\pi_{\theta}(\cdot\mid x_{\delta},Y^{(t)})
\right).
\label{eq:semantic_kl}
\end{equation}
This KL term penalizes shifts from the clean target distribution to the adversarial one, preventing the attack from achieving rejection by broadly corrupting the target model's next-token preferences.

Together, Eq.~\eqref{eq:rejection_loss} and Eq.~\eqref{eq:semantic_kl} instantiate the constrained objective in Eq.~\eqref{eq:attack_objective}. 
\(\mathcal{L}_{\mathrm{rej}}\) enlarges the information mismatch between drafter proposals and the target verifier, while \(\mathcal{L}_{\mathrm{sem}}\) limits information drift within the target model itself. 
However, maximizing \(\mathcal{L}_{\mathrm{rej}}\) under the constraint of \(\mathcal{L}_{\mathrm{sem}}\) introduces an inherent optimization conflict. 
Because the drafter is trained or designed to approximate the target verifier, gradients that suppress draft-token probability can overlap with directions that preserve the target distribution. 
A direct weighted combination can therefore be unstable: the rejection gradient may increase semantic drift, while the preservation gradient may weaken the attack signal. 
This geometric entanglement motivates the null-space projected optimization introduced next.

\subsection{Null-Space Projected Optimization}
\label{sec:null_space}

To solve the constrained problem in Eq.~\eqref{eq:attack_objective}, we seek update directions that increase rejection pressure while remaining locally feasible under the semantic-preservation constraint. 
Since the suffix \(\delta\) is discrete, we compute gradients in a continuous relaxation of the suffix, such as the embedding or one-hot token space, and use them only to score token substitutions. 
Let \(\mathbf{z}\) denote this relaxed suffix representation, and let 
\(g_{\mathrm{rej}}=\nabla_{\mathbf{z}}\mathcal{L}_{\mathrm{rej}}(x,\delta)\) and 
\(g_{\mathrm{sem}}=\nabla_{\mathbf{z}}\mathcal{L}_{\mathrm{sem}}(x,\delta)\) denote the rejection and semantic-drift gradients.

\paragraph{Local semantic null space.}
The key idea is to optimize rejection only within directions that locally preserve the target distribution. 
Under a first-order approximation, the semantic-preservation constraint defines a local feasible subspace around the current suffix. 
Let \(J_{\mathrm{sem}}(\mathbf{z})=\nabla_{\mathbf{z}}\mathcal{L}_{\mathrm{sem}}(x,\delta)^{\top}\) denote the Jacobian of the semantic-preservation objective. 
The local semantic null space is:
\begin{equation}
\mathcal{N}_{\mathrm{sem}}(\mathbf{z})
=
\left\{
\Delta
\mid
J_{\mathrm{sem}}(\mathbf{z})\Delta=0
\right\}.
\label{eq:semantic_null_space}
\end{equation}
Directions in \(\mathcal{N}_{\mathrm{sem}}(\mathbf{z})\) keep \(\mathcal{L}_{\mathrm{sem}}\) unchanged to first order, and therefore provide a local feasible subspace for rejection optimization.

\paragraph{Null-space projection.}
We construct the orthogonal projector onto the local semantic null space:
\begin{equation}
\mathbf{P}_{\mathcal{N}}
=
\mathbf{I}
-
J_{\mathrm{sem}}^{\top}
\left(
J_{\mathrm{sem}}J_{\mathrm{sem}}^{\top}
+
\xi \mathbf{I}
\right)^{-1}
J_{\mathrm{sem}},
\label{eq:null_projector}
\end{equation}
where \(\xi\) is a small damping term for numerical stability. 
The null-space rejection direction is then obtained by
\(g_{\mathrm{rej}}^{\mathcal{N}}=\mathbf{P}_{\mathcal{N}}g_{\mathrm{rej}}\). 
In our scalar KL-constraint setting, this projection reduces to an efficient rank-one operation that removes the component of \(g_{\mathrm{rej}}\) along \(g_{\mathrm{sem}}\). 
Unlike direct weighted optimization, this explicitly restricts rejection optimization to the local null space of the semantic-preservation constraint.

The final scoring direction combines feasibility restoration with null-space rejection:
\begin{equation}
g_{\mathrm{final}}
=
-
g_{\mathrm{sem}}
+
\lambda
g_{\mathrm{rej}}^{\mathcal{N}},
\label{eq:final_direction}
\end{equation}
where \(\lambda\) controls the strength of acceleration degradation. 
The first term pulls the adversarial target distribution back toward the clean target distribution, while the second term increases rejection pressure within the local semantic null space.

\paragraph{Discrete suffix update with KL-bound veto.}
The null-space direction provides a local continuous search signal, but the suffix itself consists of discrete tokens. 
We therefore use \(g_{\mathrm{final}}\) to score token substitutions and construct a candidate set \(\mathcal{C}(\delta)\) following gradient-guided suffix search. 
Since a discrete substitution may deviate from the local first-order approximation, each candidate \(\delta'\) is re-evaluated by a forward pass before selection. 
To enforce the semantic-preservation constraint in Eq.~\eqref{eq:attack_objective}, we apply a KL-bound veto and select
\begin{equation}
\delta^{\star}
=
\arg\max_{\delta' \in \mathcal{C}(\delta)}
\mathcal{L}_{\mathrm{rej}}(x,\delta')
\quad
\mathrm{s.t.}
\quad
\mathcal{L}_{\mathrm{sem}}(x,\delta') \le \epsilon .
\label{eq:kl_veto}
\end{equation}
Candidates with \(\mathcal{L}_{\mathrm{sem}}(x,\delta')>\epsilon\) are discarded, and among the remaining feasible candidates, we choose the one with the largest rejection objective. 
The suffix is then updated as \(\delta \leftarrow \delta^{\star}\). 
Overall, the null-space projection guides the local search direction, while the KL-bound veto enforces the semantic constraint on actual discrete candidates, bridging continuous optimization and discrete suffix updates.
The complete optimization procedure is summarized in Appendix~\ref{app:pseudocode}.

\begin{table}[t]
\centering
\caption{
Attack results of \textsc{Mistletoe} across models, decoding methods, and datasets.
Each entry is formatted as clean/attacked, with attacked values highlighted in red.
The final row reports the average absolute reduction marked by \(\downarrow\), with the average relative reduction shown in parentheses.
Lower attacked speed-up and accepted token length \(\tau\) indicate stronger acceleration collapse.
}
\label{tab:main_results}

\footnotesize
\renewcommand{\arraystretch}{0.95}
\setlength{\tabcolsep}{2.9pt}

\begin{tabular*}{\linewidth}{@{\extracolsep{\fill}}ll cc cc cc}
\toprule

\multirow{2}{*}{\textbf{Model}}
& \multirow{2}{*}{\textbf{Method}}
& \multicolumn{2}{c}{\textbf{MT-bench}}
& \multicolumn{2}{c}{\textbf{HumanEval}}
& \multicolumn{2}{c}{\textbf{GSM8K}} \\

\cmidrule(lr){3-4}
\cmidrule(lr){5-6}
\cmidrule(lr){7-8}

&
& Speed-up $\downarrow$ & $\tau \downarrow$
& Speed-up $\downarrow$ & $\tau \downarrow$
& Speed-up $\downarrow$ & $\tau \downarrow$ \\

\midrule

\multirow{5}{*}{V 13B}
& Medusa
& \ca{3.26$\times$}{1.48$\times$} & \ca{2.48}{2.09}
& \ca{3.42$\times$}{1.42$\times$} & \ca{2.83}{2.16}
& \ca{3.39$\times$}{1.46$\times$} & \ca{2.51}{2.32} \\
& Hydra
& \ca{4.09$\times$}{2.04$\times$} & \ca{3.49}{3.06}
& \ca{4.20$\times$}{2.03$\times$} & \ca{3.80}{3.10}
& \ca{4.29$\times$}{1.72$\times$} & \ca{3.58}{3.07} \\
& EAGLE
& \ca{3.16$\times$}{1.67$\times$} & \ca{3.37}{2.48}
& \ca{3.86$\times$}{1.79$\times$} & \ca{4.32}{3.95}
& \ca{3.66$\times$}{1.69$\times$} & \ca{3.79}{2.96} \\
& EAGLE-2
& \ca{3.95$\times$}{1.60$\times$} & \ca{4.16}{2.92}
& \ca{5.06$\times$}{2.41$\times$} & \ca{5.42}{3.55}
& \ca{4.84$\times$}{2.62$\times$} & \ca{4.91}{3.25} \\
& EAGLE-3
& \ca{5.18$\times$}{2.38$\times$} & \ca{5.73}{3.75}
& \ca{6.17$\times$}{2.77$\times$} & \ca{7.08}{3.99}
& \ca{5.47$\times$}{1.83$\times$} & \ca{5.95}{2.79} \\

\midrule

\multirow{4}{*}{V 7B}
& Medusa
& \ca{3.44$\times$}{2.38$\times$} & \ca{2.41}{1.93}
& \ca{3.75$\times$}{2.77$\times$} & \ca{2.69}{2.16}
& \ca{3.46$\times$}{2.50$\times$} & \ca{2.47}{1.82} \\
& Hydra
& \ca{4.73$\times$}{1.28$\times$} & \ca{3.54}{2.66}
& \ca{5.01$\times$}{2.08$\times$} & \ca{3.52}{2.70}
& \ca{4.37$\times$}{2.24$\times$} & \ca{3.84}{3.07} \\
& EAGLE
& \ca{2.56$\times$}{1.67$\times$} & \ca{3.18}{2.33}
& \ca{3.96$\times$}{2.10$\times$} & \ca{3.92}{3.19}
& \ca{4.11$\times$}{2.02$\times$} & \ca{3.62}{3.06} \\
& EAGLE-2
& \ca{2.54$\times$}{1.38$\times$} & \ca{4.28}{2.54}
& \ca{4.15$\times$}{3.16$\times$} & \ca{5.17}{3.03}
& \ca{4.99$\times$}{2.71$\times$} & \ca{4.68}{2.84} \\

\midrule

\textbf{Avg. drop} 
& \multicolumn{1}{c}{\textbf{--}} 
& \avgdrop{\(\downarrow\)1.89$\times$}{51.7} & \avgdrop{\(\downarrow\)0.99}{27.2} 
& \avgdrop{\(\downarrow\)2.12$\times$}{48.1} & \avgdrop{\(\downarrow\)1.21}{28.2} 
& \avgdrop{\(\downarrow\)2.20$\times$}{51.3} & \avgdrop{\(\downarrow\)1.13}{28.8} \\
\bottomrule
\end{tabular*}
\end{table}

\section{Experiments}
\label{sec:experiments}

\subsection{Experimental Setup}
\label{sec:exp_setup}

\paragraph{Models and speculative decoding systems.}
We evaluate \textsc{Mistletoe} on Vicuna-7B and Vicuna-13B target models~\citep{chiang2023vicuna}. 
For speculative decoding, we consider Medusa~\citep{cai2024medusa}, Hydra~\citep{ankner2024hydra}, EAGLE~\citep{li2024eagle}, EAGLE-2~\citep{li2024eagle2}, and EAGLE-3~\citep{li2025eagle}. 
EAGLE-3 is included only for Vicuna-13B, as its Vicuna-7B checkpoint is unavailable. 
For each setting, both the target model \(M_{\theta}\) and the drafter \(D_{\phi}\) remain fixed; the adversary only optimizes a short discrete suffix appended to the input prompt.

\paragraph{Datasets.}
We evaluate on three representative benchmarks covering open-ended dialogue, code generation, and mathematical reasoning. 
Specifically, we use all 80 questions from MT-Bench~\citep{zheng2023judging}, randomly sample 100 examples from HumanEval~\citep{chen2021evaluating}, and randomly sample 100 examples from GSM8K~\citep{cobbe2021training}. 
These datasets allow us to test \textsc{Mistletoe} across diverse generation scenarios, including instruction following, functional program synthesis, and multi-step mathematical reasoning.

\paragraph{Evaluation metrics.}
We report average accepted length \(\tau\) and speedup over vanilla autoregressive decoding as the primary efficiency metrics. 
Lower \(\tau\) indicates that fewer draft tokens are committed per target-model forward pass, while lower speedup indicates a weaker realized acceleration benefit. 
Together, they directly measure whether \textsc{Mistletoe} collapses the draft-then-verify acceleration pathway at both the acceptance and end-to-end efficiency levels.

\paragraph{Implementation details.}
All experiments are conducted on NVIDIA H20 GPUs. 
\textsc{Mistletoe} optimizes a discrete suffix \(\delta \in \mathcal{V}^{m}\) with \(m=20\), while all model parameters remain frozen. 
Clean and attacked inputs are evaluated under the same speculative decoding configuration for fair comparison. 
Additional implementation details, including optimization hyperparameters and KL bounds, are provided in Appendix~\ref{app:more_exp_details}.

\subsection{Main Results}
\label{sec:main_results}

Table~\ref{tab:main_results} presents the main effectiveness of \textsc{Mistletoe} across two Vicuna backbones, multiple speculative decoding methods, and three datasets. 
\textsc{Mistletoe} consistently reduces both speed-up and average accepted length \(\tau\) in all evaluated settings. 
On MT-Bench, the average speed-up drops by \(1.89\times\), corresponding to a \(51.7\%\) relative reduction, while \(\tau\) decreases by \(0.99\) on average. 
Similar reductions are observed on HumanEval and GSM8K, where speed-up drops by \(2.12\times\) and \(2.20\times\), and \(\tau\) decreases by \(1.21\) and \(1.13\), respectively. 
These results show that \textsc{Mistletoe} generalizes across open-ended dialogue, code generation, and mathematical reasoning.

The attack is particularly pronounced in settings with strong clean acceleration, where more accepted tokens can be disrupted. 
For example, Vicuna-13B with EAGLE-3 achieves \(6.17\times\) speed-up on HumanEval and \(5.47\times\) on GSM8K under clean decoding, but drops to \(2.77\times\) and \(1.83\times\) under \textsc{Mistletoe}. 
The corresponding accepted length also decreases from \(7.08\) to \(3.99\) and from \(5.95\) to \(2.79\), respectively. 
Even for methods with relatively lower clean acceleration, the attack remains effective. 
For instance, Medusa on Vicuna-13B drops from \(3.26\times\) to \(1.48\times\) on MT-Bench and from \(3.42\times\) to \(1.42\times\) on HumanEval, while Medusa on Vicuna-7B also decreases from \(3.44\times\) to \(2.38\times\) on MT-Bench and from \(3.75\times\) to \(2.77\times\) on HumanEval.

Overall, the consistent decrease in \(\tau\) confirms that the loss of speed-up is driven by acceptance collapse. 
Since \(\tau\) measures how many tokens are committed per target-model forward pass, its reduction indicates that fewer drafter-proposed tokens survive verification. 
These results further support our central mechanism-level claim that \textsc{Mistletoe} disables speculative acceleration by degrading draft-token acceptability across models, methods, and tasks.

\begin{figure*}[t]
    \centering
    \includegraphics[width=\textwidth]{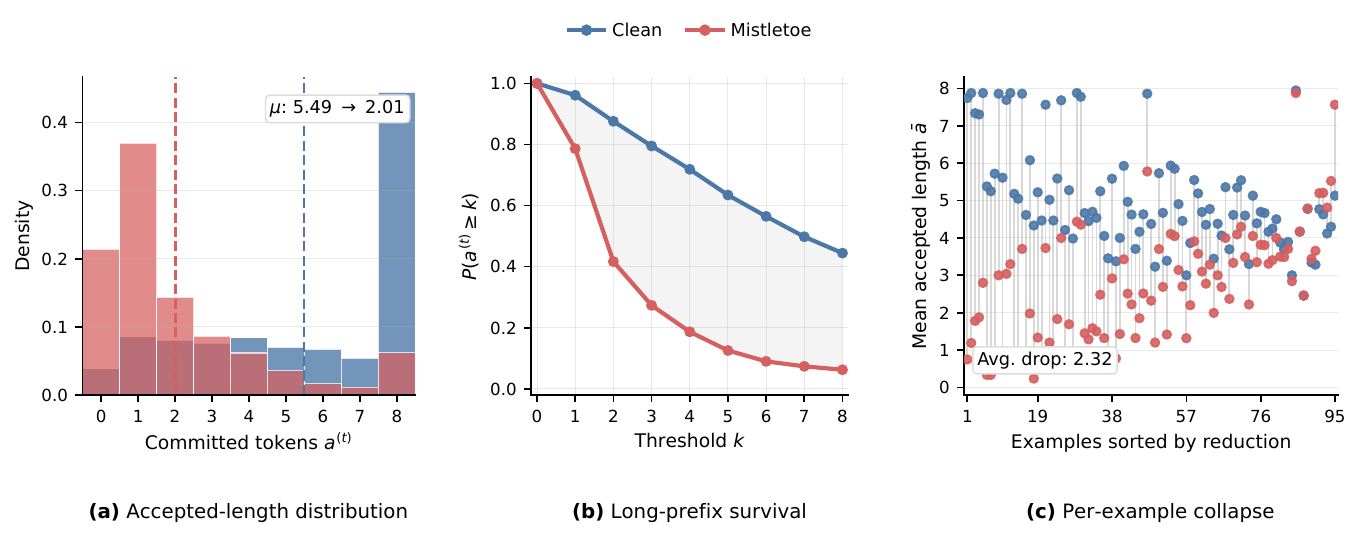}
    \caption{
    Mechanism visualization of acceptance collapse.
    The figure compares clean speculative decoding and \textsc{Mistletoe} using verification-cycle-level accepted lengths.
    \textbf{(a) Accepted-length distribution:} \textsc{Mistletoe} shifts the distribution of committed tokens per target-model forward pass \(a^{(t)}\) toward smaller values.
    \textbf{(b) Long-prefix survival:} the survival probability \(P(a^{(t)} \ge k)\) drops sharply under attack, showing that long draft prefixes rarely survive verification.
    \textbf{(c) Per-example collapse:} the mean accepted length decreases across most prompts, indicating that the collapse is broadly observed rather than driven by isolated outliers.
    }
    \label{fig:acceptance_collapse}
\end{figure*}

Figure~\ref{fig:acceptance_collapse} further explains the source of speed-up degradation.
Compared with clean speculative decoding, \textsc{Mistletoe} shifts the distribution of committed tokens per target-model forward pass \(a^{(t)}\) toward smaller values and sharply reduces the survival probability \(P(a^{(t)}\ge k)\) for long accepted prefixes.
The per-example view shows that this reduction is broadly observed across prompts rather than caused by isolated outliers.
These observations align with the \(\tau\) reductions in Table~\ref{tab:main_results}, confirming that the attack disables speculative acceleration by collapsing draft-token acceptance.

\begin{table}[t]
\centering
\caption{
Ablation study of \textsc{Mistletoe} components.
Lower speed-up and \(\tau\) indicate stronger acceleration collapse; lower PPL and Rep-4 indicate more natural and less repetitive outputs.
}
\label{tab:ablation}

\footnotesize
\renewcommand{\arraystretch}{1.05}
\setlength{\tabcolsep}{4.5pt}

\begin{tabular*}{\linewidth}{@{\extracolsep{\fill}}lccc cccc}
\toprule

\multirow{2}{*}{\textbf{Configuration}}
& \multicolumn{3}{c}{\textbf{Component}}
& \multicolumn{2}{c}{\textbf{Acceleration}}
& \multicolumn{2}{c}{\textbf{Output Normality}} \\

\cmidrule(lr){2-4}
\cmidrule(lr){5-6}
\cmidrule(lr){7-8}

& \(\mathcal{L}_{\mathrm{rej}}\)
& \(\mathcal{L}_{\mathrm{sem}}\)
& Projected Opt.
& Speed-up \(\downarrow\)
& \(\tau \downarrow\)
& PPL \(\downarrow\)
& Rep-4 \(\downarrow\) \\

\midrule

Clean
& \xmark & \xmark & \xmark
& \(5.47\times\) & 5.95 & \textbf{2.5} & 0.1813 \\

\(\mathcal{L}_{\mathrm{rej}}\) only
& \cmark & \xmark & \xmark
& \(3.30\times\) & 3.41 & 334.1 & 0.0844 \\

\(\mathcal{L}_{\mathrm{sem}}\) only
& \xmark & \cmark & \xmark
& \(4.47\times\) & 4.73 & 213.2 & 0.0634 \\

Naive joint
& \cmark & \cmark & \xmark
& \(3.73\times\) & 4.13 & 196.6 & 0.0952 \\

\textbf{Full \textsc{Mistletoe}}
& \cmark & \cmark & \cmark
& \(\mathbf{1.83}\times\) & \(\mathbf{2.79}\) & 49.2 & \(\mathbf{0.0111}\) \\

\bottomrule
\end{tabular*}
\end{table}

\subsection{Ablation Study}
\label{sec:ablation}

Table~\ref{tab:ablation} evaluates the contribution of each component in \textsc{Mistletoe}. 
Using only \(\mathcal{L}_{\mathrm{rej}}\) reduces speed-up from \(5.47\times\) to \(3.30\times\) and decreases \(\tau\) from \(5.95\) to \(3.41\), confirming that target-side draft-token surprisal provides a direct signal for degrading the acceptance pathway. 
However, this variant produces extremely high PPL, indicating that aggressive rejection optimization alone can lead to abnormal generations. 
Using only \(\mathcal{L}_{\mathrm{sem}}\) is more conservative but much weaker in acceleration degradation, reducing speed-up only to \(4.47\times\).

Naively combining \(\mathcal{L}_{\mathrm{rej}}\) and \(\mathcal{L}_{\mathrm{sem}}\) does not fully resolve this trade-off. 
Although it incorporates both objectives, it still yields high PPL and weaker acceleration collapse than the full method, suggesting that direct joint optimization struggles to coordinate rejection pressure and output normality. 
In contrast, full \textsc{Mistletoe} achieves the strongest acceleration collapse, reducing speed-up to \(1.83\times\) and \(\tau\) to \(2.79\), while substantially lowering PPL and Rep-4 compared with naive objectives. 
These results show that projected optimization is crucial for balancing the rejection and preservation objectives, enabling \textsc{Mistletoe} to degrade speculative acceleration without relying on visibly unnatural or repetitive outputs.

\subsection{Transferability Analysis}
\label{sec:transferability}

Table~\ref{tab:transfer_results} evaluates whether adversarial suffixes optimized on EAGLE-3 transfer to other speculative decoding methods. 
Across all target methods and datasets, transferred suffixes consistently reduce both speed-up and accepted length \(\tau\), indicating that \textsc{Mistletoe} captures a cross-method vulnerability rather than overfitting to the source decoding method. 
The transfer is especially strong on Medusa: speed-up drops from \(3.68\times\) to \(1.03\times\) on MT-Bench, from \(3.11\times\) to \(1.15\times\) on HumanEval, and from \(3.39\times\) to \(1.06\times\) on GSM8K. 
This shows that suffixes generated on EAGLE-3 can substantially disrupt even a different speculative decoding pipeline.

The transferred attack also remains effective on Hydra, EAGLE, and EAGLE-2, although the magnitude varies across methods and datasets. 
For example, EAGLE-2 drops from \(4.44\times\) to \(2.03\times\) on MT-Bench and from \(5.00\times\) to \(2.29\times\) on HumanEval, while Hydra drops from \(4.59\times\) to \(2.33\times\) on MT-Bench and from \(4.26\times\) to \(1.91\times\) on HumanEval. 
The consistent decrease in \(\tau\) further suggests that transferability arises from degraded draft-token acceptance, supporting our claim that \textsc{Mistletoe} exploits a shared drafter--target agreement dependency across speculative decoding methods.

\begin{table}[t]
\centering
\caption{
Cross-method transferability of adversarial suffixes generated on EAGLE-3.
Each entry is formatted as clean/transferred, with transferred values highlighted in red.
Lower transferred speed-up and \(\tau\) indicate stronger transferability.
}
\label{tab:transfer_results}

\footnotesize
\renewcommand{\arraystretch}{0.95}
\setlength{\tabcolsep}{3.2pt}

\begin{tabular*}{\linewidth}{@{\extracolsep{\fill}}l cc cc cc}
\toprule

\multirow{2}{*}{\textbf{Target Method}}
& \multicolumn{2}{c}{\textbf{MT-Bench}}
& \multicolumn{2}{c}{\textbf{HumanEval}}
& \multicolumn{2}{c}{\textbf{GSM8K}} \\

\cmidrule(lr){2-3}
\cmidrule(lr){4-5}
\cmidrule(lr){6-7}

& Speed-up $\downarrow$ & $\tau \downarrow$
& Speed-up $\downarrow$ & $\tau \downarrow$
& Speed-up $\downarrow$ & $\tau \downarrow$ \\

\midrule

Medusa
& \ca{3.68$\times$}{1.03$\times$} & \ca{2.48}{1.56}
& \ca{3.11$\times$}{1.15$\times$} & \ca{2.83}{1.81}
& \ca{3.39$\times$}{1.06$\times$} & \ca{2.51}{1.67} \\

Hydra
& \ca{4.59$\times$}{2.33$\times$} & \ca{3.49}{2.58}
& \ca{4.26$\times$}{1.91$\times$} & \ca{3.80}{2.74}
& \ca{4.29$\times$}{1.95$\times$} & \ca{3.58}{3.19} \\

EAGLE
& \ca{3.57$\times$}{1.97$\times$} & \ca{3.38}{2.61}
& \ca{3.90$\times$}{1.59$\times$} & \ca{4.32}{3.46}
& \ca{3.67$\times$}{1.63$\times$} & \ca{3.79}{2.88} \\

EAGLE-2
& \ca{4.44$\times$}{2.03$\times$} & \ca{4.16}{3.03}
& \ca{5.00$\times$}{2.29$\times$} & \ca{5.42}{3.75}
& \ca{4.84$\times$}{2.41$\times$} & \ca{4.91}{3.33} \\

\bottomrule
\end{tabular*}
\end{table}

\section{Conclusion}
\label{sec:conclusion}

We introduced \textsc{Mistletoe}, an acceleration-collapse attack against speculative decoding. 
Instead of corrupting final responses, it reduces drafter-token acceptability under the target verifier through target-side surprisal, KL-bounded preservation, and null-space projected optimization. 
Experiments across Vicuna backbones, decoding methods, and benchmarks show consistent reductions in speed-up and \(\tau\). 
Ablations and transfer results validate the design and reveal a shared vulnerability in speculative decoding, motivating more robust acceleration mechanisms.


\bibliographystyle{plainnat}
\bibliography{main}

\begin{thebibliography}{20}
\providecommand{\natexlab}[1]{#1}
\providecommand{\url}[1]{\texttt{#1}}
\expandafter\ifx\csname urlstyle\endcsname\relax
  \providecommand{\doi}[1]{doi: #1}\else
  \providecommand{\doi}{doi: \begingroup \urlstyle{rm}\Url}\fi

\bibitem[Ankner et~al.(2024)Ankner, Parthasarathy, Nrusimha, Rinard, Ragan-Kelley, and Brandon]{ankner2024hydra}
Zachary Ankner, Rishab Parthasarathy, Aniruddha Nrusimha, Christopher Rinard, Jonathan Ragan-Kelley, and William Brandon.
\newblock Hydra: Sequentially-dependent draft heads for medusa decoding.
\newblock \emph{arXiv preprint arXiv:2402.05109}, 2024.

\bibitem[Cai et~al.(2024)Cai, Li, Geng, Peng, Lee, Chen, and Dao]{cai2024medusa}
Tianle Cai, Yuhong Li, Zhengyang Geng, Hongwu Peng, Jason~D Lee, Deming Chen, and Tri Dao.
\newblock Medusa: Simple llm inference acceleration framework with multiple decoding heads.
\newblock \emph{arXiv preprint arXiv:2401.10774}, 2024.

\bibitem[Chen et~al.(2023)Chen, Borgeaud, Irving, Lespiau, Sifre, and Jumper]{chen2023accelerating}
Charlie Chen, Sebastian Borgeaud, Geoffrey Irving, Jean-Baptiste Lespiau, Laurent Sifre, and John Jumper.
\newblock Accelerating large language model decoding with speculative sampling.
\newblock \emph{arXiv preprint arXiv:2302.01318}, 2023.

\bibitem[Chen et~al.(2021)Chen, Tworek, Jun, Yuan, Pinto, Kaplan, Edwards, Burda, Joseph, Brockman, et~al.]{chen2021evaluating}
Mark Chen, Jerry Tworek, Heewoo Jun, Qiming Yuan, Henrique Ponde De~Oliveira Pinto, Jared Kaplan, Harri Edwards, Yuri Burda, Nicholas Joseph, Greg Brockman, et~al.
\newblock Evaluating large language models trained on code.
\newblock \emph{arXiv preprint arXiv:2107.03374}, 2021.

\bibitem[Chiang et~al.(2023)Chiang, Li, Lin, Sheng, Wu, Zhang, Zheng, Zhuang, Zhuang, Gonzalez, et~al.]{chiang2023vicuna}
Wei-Lin Chiang, Zhuohan Li, Ziqing Lin, Ying Sheng, Zhanghao Wu, Hao Zhang, Lianmin Zheng, Siyuan Zhuang, Yonghao Zhuang, Joseph~E Gonzalez, et~al.
\newblock Vicuna: An open-source chatbot impressing gpt-4 with 90\%* chatgpt quality.
\newblock \emph{See https://vicuna. lmsys. org (accessed 14 April 2023)}, 2\penalty0 (3):\penalty0 6, 2023.

\bibitem[Cobbe et~al.(2021)Cobbe, Kosaraju, Bavarian, Chen, Jun, Kaiser, Plappert, Tworek, Hilton, Nakano, et~al.]{cobbe2021training}
Karl Cobbe, Vineet Kosaraju, Mohammad Bavarian, Mark Chen, Heewoo Jun, Lukasz Kaiser, Matthias Plappert, Jerry Tworek, Jacob Hilton, Reiichiro Nakano, et~al.
\newblock Training verifiers to solve math word problems.
\newblock \emph{arXiv preprint arXiv:2110.14168}, 2021.

\bibitem[Fang et~al.(2024)Fang, Jiang, Wang, Ma, Jie, Wang, He, and Chua]{fang2024alphaedit}
Junfeng Fang, Houcheng Jiang, Kun Wang, Yunshan Ma, Shi Jie, Xiang Wang, Xiangnan He, and Tat-Seng Chua.
\newblock Alphaedit: Null-space constrained knowledge editing for language models.
\newblock \emph{arXiv preprint arXiv:2410.02355}, 2024.

\bibitem[Grattafiori et~al.(2024)Grattafiori, Dubey, Jauhri, Pandey, Kadian, Al-Dahle, Letman, Mathur, Schelten, Vaughan, et~al.]{grattafiori2024llama}
Aaron Grattafiori, Abhimanyu Dubey, Abhinav Jauhri, Abhinav Pandey, Abhishek Kadian, Ahmad Al-Dahle, Aiesha Letman, Akhil Mathur, Alan Schelten, Alex Vaughan, et~al.
\newblock The llama 3 herd of models.
\newblock \emph{arXiv preprint arXiv:2407.21783}, 2024.

\bibitem[Hu et~al.(2025{\natexlab{a}})Hu, Li, Xie, Lu, Toh, and Zhou]{hu2025griffin}
Shijing Hu, Jingyang Li, Xingyu Xie, Zhihui Lu, Kim-Chuan Toh, and Pan Zhou.
\newblock Griffin: Effective token alignment for faster speculative decoding.
\newblock \emph{arXiv preprint arXiv:2502.11018}, 2025{\natexlab{a}}.

\bibitem[Hu et~al.(2025{\natexlab{b}})Hu, Liu, Dong, Peng, McDanel, and Zhang]{hu2025speculative}
Yunhai Hu, Zining Liu, Zhenyuan Dong, Tianfan Peng, Bradley McDanel, and Sai~Qian Zhang.
\newblock Speculative decoding and beyond: An in-depth survey of techniques.
\newblock \emph{arXiv preprint arXiv:2502.19732}, 2025{\natexlab{b}}.

\bibitem[Leviathan et~al.(2023)Leviathan, Kalman, and Matias]{leviathan2023fast}
Yaniv Leviathan, Matan Kalman, and Yossi Matias.
\newblock Fast inference from transformers via speculative decoding.
\newblock In \emph{International Conference on Machine Learning}, pages 19274--19286. PMLR, 2023.

\bibitem[Li et~al.(2024{\natexlab{a}})Li, Wei, Zhang, and Zhang]{li2024eagle}
Yuhui Li, Fangyun Wei, Chao Zhang, and Hongyang Zhang.
\newblock Eagle: Speculative sampling requires rethinking feature uncertainty.
\newblock \emph{arXiv preprint arXiv:2401.15077}, 2024{\natexlab{a}}.

\bibitem[Li et~al.(2024{\natexlab{b}})Li, Wei, Zhang, and Zhang]{li2024eagle2}
Yuhui Li, Fangyun Wei, Chao Zhang, and Hongyang Zhang.
\newblock Eagle-2: Faster inference of language models with dynamic draft trees.
\newblock In \emph{Proceedings of the 2024 conference on empirical methods in natural language processing}, pages 7421--7432, 2024{\natexlab{b}}.

\bibitem[Li et~al.(2025)Li, Wei, Zhang, and Zhang]{li2025eagle}
Yuhui Li, Fangyun Wei, Chao Zhang, and Hongyang Zhang.
\newblock Eagle-3: Scaling up inference acceleration of large language models via training-time test.
\newblock \emph{arXiv preprint arXiv:2503.01840}, 2025.

\bibitem[Liu et~al.(2024)Liu, Feng, Xue, Wang, Wu, Lu, Zhao, Deng, Zhang, Ruan, et~al.]{liu2024deepseek}
Aixin Liu, Bei Feng, Bing Xue, Bingxuan Wang, Bochao Wu, Chengda Lu, Chenggang Zhao, Chengqi Deng, Chenyu Zhang, Chong Ruan, et~al.
\newblock Deepseek-v3 technical report.
\newblock \emph{arXiv preprint arXiv:2412.19437}, 2024.

\bibitem[Wang et~al.(2025{\natexlab{a}})Wang, Liu, Hu, Wu, and He]{wang2025secdecoding}
Jiayou Wang, Rundong Liu, Yue Hu, Huijia Wu, and Zhaofeng He.
\newblock Secdecoding: Steerable decoding for safer llm generation.
\newblock In \emph{Findings of the Association for Computational Linguistics: EMNLP 2025}, pages 20504--20521, 2025{\natexlab{a}}.

\bibitem[Wang et~al.(2025{\natexlab{b}})Wang, Zhu, and Cheng]{wang2025speculative}
Xuekang Wang, Shengyu Zhu, and Xueqi Cheng.
\newblock Speculative safety-aware decoding.
\newblock In \emph{Proceedings of the 2025 Conference on Empirical Methods in Natural Language Processing}, pages 12838--12852, 2025{\natexlab{b}}.

\bibitem[Wei et~al.(2024)Wei, Abdulrazzag, Zhang, Muursepp, and Saileshwar]{wei2024speculation}
Jiankun Wei, Abdulrahman Abdulrazzag, Tianchen Zhang, Adel Muursepp, and Gururaj Saileshwar.
\newblock When speculation spills secrets: Side channels via speculative decoding in llms.
\newblock \emph{arXiv preprint arXiv:2411.01076}, 2024.

\bibitem[Yang et~al.(2025)Yang, Li, Yang, Zhang, Hui, Zheng, Yu, Gao, Huang, Lv, et~al.]{yang2025qwen3}
An~Yang, Anfeng Li, Baosong Yang, Beichen Zhang, Binyuan Hui, Bo~Zheng, Bowen Yu, Chang Gao, Chengen Huang, Chenxu Lv, et~al.
\newblock Qwen3 technical report.
\newblock \emph{arXiv preprint arXiv:2505.09388}, 2025.

\bibitem[Zheng et~al.(2023)Zheng, Chiang, Sheng, Zhuang, Wu, Zhuang, Lin, Li, Li, Xing, et~al.]{zheng2023judging}
Lianmin Zheng, Wei-Lin Chiang, Ying Sheng, Siyuan Zhuang, Zhanghao Wu, Yonghao Zhuang, Zi~Lin, Zhuohan Li, Dacheng Li, Eric Xing, et~al.
\newblock Judging llm-as-a-judge with mt-bench and chatbot arena.
\newblock \emph{Advances in neural information processing systems}, 36:\penalty0 46595--46623, 2023.

\end{thebibliography}

\clearpage
\appendix

\section{More Experimental Configuration}
\label{app:more_exp_details}

We generate adversarial suffixes for text-based prompts to disrupt the efficiency of speculative decoding systems. 
We evaluate \textsc{Mistletoe} on several widely used speculative decoding frameworks and describe their implementation settings below. 
Unless otherwise specified, all systems use their standard speculative decoding configurations.

\paragraph{Speculative decoding systems.}
We evaluate the following speculative decoding systems.

\begin{itemize}
    \item \textbf{EAGLE.}
    EAGLE~\citep{li2024eagle} uses tree-based speculative decoding for feature-level drafting. 
    We adopt the standard \texttt{mc\_sim\_7b\_63} tree configuration to organize and verify candidate token sequences.

    \item \textbf{EAGLE-2.}
    EAGLE-2~\citep{li2024eagle2} extends EAGLE with an enhanced dynamic drafting tree. 
    To isolate the EAGLE-2 architecture from EAGLE-3 components, we disable EAGLE-3-specific features using \texttt{--no-eagle3}.

    \item \textbf{EAGLE-3.}
    EAGLE-3~\citep{li2025eagle} improves feature-level alignment by using multiple intermediate hidden states from the target model. 
    We enable this mechanism with \texttt{--use-eagle3}, which uses early-, middle-, and late-layer hidden states for drafting.

    \item \textbf{Medusa.}
    Medusa~\citep{cai2024medusa} adopts a multi-head parallel speculative decoding design. 
    We use its native setting with five Medusa heads, verification over up to ten draft candidates, and a KV-cache margin of 128 tokens. 
    When pre-computing clean reference logits for the KL objective, we temporarily remove the Medusa attention mask to obtain the target reference distribution; inference-time decoding follows the standard Medusa pipeline.

    \item \textbf{Hydra.}
    Hydra~\citep{ankner2024hydra} uses multiple prediction heads with a posterior acceptance mechanism. 
    We use its default posterior acceptance setting, with threshold \(0.09\) and coefficient \(0.3\). 
    As with Medusa, we disable the Hydra attention mask only when pre-computing clean reference logits for the KL objective, while keeping standard Hydra decoding during evaluation.
\end{itemize}

\paragraph{Optimization hyperparameters.}
We use the same adversarial optimization protocol across all target models and speculative decoding systems. 
The maximum number of suffix-optimization iterations is set to \(50\). 
The semantic-preservation objective is estimated over \(20\) predictive positions. 
The null-space rejection weight is fixed to \(\lambda=2.0\), corresponding to Eq.~\eqref{eq:final_direction}. 
The optimized suffix is directly appended to the clean input prompt.

\paragraph{Dataset-specific KL bounds.}
To bound target-distribution drift during discrete candidate selection, we use dataset-specific KL thresholds. 
The threshold is set to \(5.0\) for GSM8K~\citep{cobbe2021training}, \(7.0\) for MT-Bench~\citep{zheng2023judging}, and \(15.0\) for HumanEval~\citep{chen2021evaluating}. 
These values define the maximum allowable semantic-preservation loss \(\mathcal{L}_{\mathrm{sem}}\) during the KL-bound veto.

\paragraph{General implementation protocol.}
All experiments are conducted using FP16 precision on NVIDIA H20 GPUs. 
All target LLMs and drafters are evaluated in their standard inference modes without task-specific fine-tuning. 
We use greedy decoding with temperature \(0.0\) to reduce sampling randomness and isolate efficiency degradation caused by adversarial suffixes. 
The maximum generation length is capped at \(512\) tokens unless otherwise specified.

\paragraph{Metric computation.}
Speed-up and average accepted length \(\tau\) are measured under the same decoding configuration for clean and attacked prompts. 
All reported metrics are averaged over multiple independent inference runs to reduce system-level variance. 
Clean and attacked settings use identical model weights, drafter configurations, decoding parameters, and evaluation prompts; the only difference is whether the optimized adversarial suffix is appended.

\paragraph{Filtering abnormal empty outputs.}
In rare cases, speculative decoding implementations may return abnormal empty outputs due to decoding or runtime edge cases unrelated to the attack objective. 
Such cases arise from the speculative decoding implementation itself and are not specific to \textsc{Mistletoe} or its optimization objective. 
We exclude such invalid runs from metric aggregation and apply the same rule to clean and attacked settings. 
This filtering only removes empty outputs and does not filter based on speed-up, accepted length, perplexity, repetition, or response quality.

\section{Algorithm Pseudocode}
\label{app:pseudocode}

Algorithm~\ref{alg:mistletoe} summarizes the optimization procedure of \textsc{Mistletoe}. 
The attack optimizes a discrete suffix \(\delta\) while keeping the target model \(M_{\theta}\) and drafter \(D_{\phi}\) fixed. 
At each iteration, \textsc{Mistletoe} computes the rejection gradient and the semantic-preservation gradient in a continuous relaxation of the suffix, projects the rejection direction onto the local semantic null space, and uses the resulting direction to guide discrete suffix search. 
Since the projection is only a local first-order approximation, candidate suffixes are further evaluated by a forward pass and filtered by a KL-bound veto.

\begin{algorithm}[h]
\caption{\textsc{Mistletoe}: Null-Space Guided Suffix Optimization}
\label{alg:mistletoe}
\begin{algorithmic}[1]
\REQUIRE Prompt \(x\), target model \(M_{\theta}\), drafter \(D_{\phi}\), iterations \(T\), suffix length \(m\), candidate size \(K\), evaluation batch size \(B\), KL bound \(\epsilon\), rejection weight \(\lambda\), damping coefficient \(\xi\)
\ENSURE Adversarial suffix \(\delta^{\star}\)

\STATE Initialize a discrete suffix \(\delta \in \mathcal{V}^{m}\)

\FOR{\(r = 1\) \TO \(T\)}
    \STATE \(x_{\delta} \gets x \oplus \delta\)
    
    \STATE Run speculative decoding with \(D_{\phi}\) and \(M_{\theta}\) on \(x_{\delta}\) to obtain drafter-proposed tokens \(\{\hat y_i^{(t)}\}_{(t,i)\in\mathcal{I}}\)
    
    \STATE Estimate the rejection objective
    \[
    \mathcal{L}_{\mathrm{rej}}(x,\delta)
    =
    \frac{1}{|\mathcal{I}|}
    \sum_{(t,i)\in\mathcal{I}}
    -\log
    \pi_{\theta}\!\left(
    \hat y_i^{(t)}
    \mid
    x_{\delta},Y^{(t)},\hat y_{<i}^{(t)}
    \right)
    \]
    
    \STATE Estimate the semantic-preservation objective
    \[
    \mathcal{L}_{\mathrm{sem}}(x,\delta)
    =
    \frac{1}{|\mathcal{S}|}
    \sum_{t\in\mathcal{S}}
    D_{\mathrm{KL}}
    \left(
    \pi_{\theta}(\cdot\mid x,Y^{(t)})
    \,\middle\|\,
    \pi_{\theta}(\cdot\mid x_{\delta},Y^{(t)})
    \right)
    \]
    
    \STATE Let \(\mathbf{z}\) be a continuous relaxation of \(\delta\)
    \STATE \(g_{\mathrm{rej}} \gets \nabla_{\mathbf{z}}\mathcal{L}_{\mathrm{rej}}(x,\delta)\)
    \STATE \(g_{\mathrm{sem}} \gets \nabla_{\mathbf{z}}\mathcal{L}_{\mathrm{sem}}(x,\delta)\)
    
    \STATE Project the rejection gradient onto the local semantic null space:
    \[
    g_{\mathrm{rej}}^{\mathcal{N}}
    \gets
    g_{\mathrm{rej}}
    -
    \frac{
    \langle g_{\mathrm{rej}}, g_{\mathrm{sem}}\rangle
    }{
    \|g_{\mathrm{sem}}\|_2^2+\xi
    }
    g_{\mathrm{sem}}
    \]
    
    \STATE Construct the final scoring direction:
    \[
    g_{\mathrm{final}}
    \gets
    -
    g_{\mathrm{sem}}
    +
    \lambda g_{\mathrm{rej}}^{\mathcal{N}}
    \]
    
    \STATE Use \(g_{\mathrm{final}}\) to propose top-\(K\) token substitutions and form a candidate set \(\mathcal{C}(\delta)\)
    
    \STATE \(\delta^{\star} \gets \delta,\quad \mathcal{L}_{\mathrm{best}} \gets -\infty\)
    
    \FOR{\(b = 1\) \TO \(B\)}
        \STATE Sample a candidate suffix \(\delta_b \in \mathcal{C}(\delta)\)
        \STATE Recompute \(\mathcal{L}_{\mathrm{rej}}(x,\delta_b)\) and \(\mathcal{L}_{\mathrm{sem}}(x,\delta_b)\) by forward evaluation
        
        \IF{\(\mathcal{L}_{\mathrm{sem}}(x,\delta_b) \le \epsilon\)}
            \IF{\(\mathcal{L}_{\mathrm{rej}}(x,\delta_b) > \mathcal{L}_{\mathrm{best}}\)}
                \STATE \(\mathcal{L}_{\mathrm{best}} \gets \mathcal{L}_{\mathrm{rej}}(x,\delta_b)\)
                \STATE \(\delta^{\star} \gets \delta_b\)
            \ENDIF
        \ENDIF
    \ENDFOR
    
    \STATE \(\delta \gets \delta^{\star}\)
\ENDFOR

\RETURN \(\delta^{\star}\)
\end{algorithmic}
\end{algorithm}

\section{Limitations and Responsible Use}
\label{app:limitations}

\paragraph{Scope of evaluation.}
Our study focuses on Vicuna-7B and Vicuna-13B with representative speculative decoding systems, including Medusa, Hydra, EAGLE, EAGLE-2, and EAGLE-3. 
This setting allows us to systematically evaluate whether adversarial suffixes can degrade the draft-verification pathway across multiple acceleration designs and generation tasks. 
While the evaluated systems cover widely used speculative decoding paradigms, future work may extend the analysis to additional model families, larger backbones, and production-serving configurations.

\paragraph{Attack setting.}
\textsc{Mistletoe} primarily uses white-box gradients during suffix construction while keeping the target model and drafter fixed. 
This setting provides a controlled way to expose mechanism-level vulnerabilities in speculative decoding and to analyze how drafter--target mismatch can be amplified. 
Importantly, our transferability experiments show that suffixes optimized on one source method can still degrade other speculative decoding methods, suggesting that the attack is not tied to a single white-box configuration. 
A promising future direction is to further study fully black-box or query-limited variants, especially for closed-source deployments where direct gradient access is unavailable.

\paragraph{Output normality evaluation.}
The attack is designed to degrade speculative acceleration without relying on visibly abnormal outputs. 
We evaluate output normality using PPL and Rep-4, and provide qualitative examples to further inspect generated responses. 
These measurements capture fluency and repetitive degeneration, but they do not exhaustively characterize all aspects of semantic equivalence or task correctness. 
Future evaluations may incorporate task-specific correctness metrics or human/LLM-based judgments to provide a more fine-grained assessment.

\paragraph{Responsible use.}
This work aims to reveal a performance-robustness risk in speculative decoding and to motivate more secure acceleration mechanisms. 
Because the proposed attack could be misused to increase serving cost or reduce the efficiency of deployed LLM systems, it should be used for research, auditing, and defense development rather than for disrupting real-world services. 
Potential defenses include monitoring accepted-length distributions, detecting abnormal drafter--target mismatch, and designing verification mechanisms that are more robust to adversarial prompt perturbations.

\section{LLM Usage}
We used an OpenAI LLM (GPT-5) as a writing
and formatting assistant. In particular, it helped
refine grammar and phrasing, improve textual flow
and clarity, and suggest edits to figure/table captions and layout (e.g., column alignment, caption
length, placement). The LLM did not contribute to
research ideation, experimental design, implementation, data analysis, or technical content beyond
surface-level edits. All outputs were carefully reviewed and edited by the authors, who take full
responsibility for the final text and visuals.

\end{document}